\documentclass[letterpaper]{article}
\usepackage{aaai}
\usepackage{times}
\usepackage{helvet}
\usepackage{courier}
\usepackage{hyperref}
\usepackage{xcolor}
\definecolor{darkblue}{RGB}{0,0,139}
\hypersetup{
    colorlinks=true,
    linkcolor=black,
    citecolor=darkblue,      
    urlcolor=darkblue,       
    filecolor=darkblue,
    breaklinks=true
}

\usepackage{amsmath}
\usepackage{amssymb}
\usepackage{mathtools}
\usepackage{amsthm}
\usepackage{graphicx}
\usepackage{enumerate}
\usepackage{xcolor}
\usepackage{natbib} 
\usepackage{url} 

\usepackage{microtype}
\usepackage{adjustbox}
\usepackage{subfigure}
\usepackage{tabularx}
\usepackage{array}
\usepackage{booktabs} 
\usepackage{multirow} 

\title{Automated Procedural Analysis via Video-Language Models for AI-assisted Nursing Skills Assessment}
\author{
    Shen Chang\textsuperscript{1}, 
    Dennis Liu\textsuperscript{2}, 
    Renran Tian\textsuperscript{3}, 
    Kristen L. Swartzell\textsuperscript{4}, 
    Stacie L. Klingler\textsuperscript{4}, 
    Amy M. Nagle\textsuperscript{1}, 
    Nan Kong\textsuperscript{1} \\
    \textsuperscript{1}Weldon School of Biomedical Engineering, Purdue University, West Lafayette, IN, USA \\
    \textsuperscript{2}Department of Industrial and Operations Engineering, University of Michigan, Ann Arbor, MI, USA \\
    \textsuperscript{3}Edward P. Fitts Department of Industrial and Systems Engineering, North Carolina State University, Raleigh, NC, USA \\
    \textsuperscript{4}School of Nursing, Purdue University, West Lafayette, IN, USA \\
    \url{https://github.com/chang887/nursing-vlm-assessment}\\
}

\begin{document}

\maketitle

\begin{abstract}
Consistent high-quality nursing care is essential for patient safety, yet current nursing education depends on subjective, time-intensive instructor feedback in training future nurses, which limits scalability and efficiency in their training, and thus hampers nursing competency when they enter the workforce. In this paper, we introduce a video-language model (VLM) based framework to develop the AI capability of automated procedural assessment and feedback for nursing skills training, with the potential of being integrated into existing training programs. Mimicking human skill acquisition, the framework follows a curriculum-inspired progression, advancing from high-level action recognition, fine-grained subaction decomposition, and ultimately to procedural reasoning. This design supports scalable evaluation by reducing instructor workload while preserving assessment quality. The system provides three core capabilities: 1) diagnosing errors by identifying missing or incorrect subactions in nursing skill instruction videos, 2) generating explainable feedback by clarifying why a step is out of order or omitted, and 3) enabling objective, consistent formative evaluation of procedures. Validation on synthesized videos demonstrates reliable error detection and temporal localization, confirming its potential to handle real-world training variability. By addressing workflow bottlenecks and supporting large-scale, standardized evaluation, this work advances AI applications in nursing education, contributing to stronger workforce development and ultimately safer patient care.
\end{abstract}

\section{Introduction}

The resilience of healthcare systems rests heavily on a well-trained and proficient nursing workforce. Nurses, constituting nearly 50\% of healthcare workers worldwide, perform safety-critical procedures at workplace that directly shape patient outcomes and overall well-being~\citep{who2025}. Yet, healthcare systems face a dual challenge: an escalating shortage of qualified nurses and growing demands for care delivery~\citep{aacn2023,aacn2024,uschamber2024}. This imbalance places unprecedented pressure on nursing education programs worldwide to expand their training capacity while upholding an uncompromising standard on the training performance. This pressure is particularly felt in the training of nursing skills. Nursing skills encompass both hard (technical) skills like patient assessment, medication administration, wound care, and cardiopulmonary resuscitation (CPR), and soft (interpersonal) skills such as communication, empathy, critical thinking, problem-solving, and adaptability. Effective nursing requires a strong combination of these skills to provide quality patient care, manage complex situations, and collaborate with healthcare teams.

Current assessment methods—both manual and automated—fall short of capturing the full complexity of nursing skills. The effective evaluation requires coordinated clinical reasoning, precise temporal execution, and integrated decision-making~\citep{mukurunge2024,ampt2024,OVREBO2025}. In recent years, the shift toward competency-based education has further intensified this assessment burden, as it requires progressive demonstrations of nursing skills competency throughout the curriculum rather than the traditional one-time evaluation~\citep{AACN2021}. These limitations create an urgent need for intelligent systems capable of evaluating procedural competence, detecting errors, and providing interpretable, actionable guidance to learners.

Emerging advances in video-language models (VLMs) offer an opportunity to bridge this analytical gap~\citep{ashutosh2025expertaf, li2022blip, survey2023video, damonlpsg2023videollama}. VLMs demonstrate a remarkable capability of analyzing complex sequential activities, understanding temporal relationships, and generating context-aware feedback. Unlike conventional action recognition models, VLMs integrate visual perception with reasoning about procedural correctness, safety, and skill progression. Despite these strengths, their application to nursing skills assessment remains largely unexplored, as the corresponding clinical tasks demand exceptional precision, fine-grained temporal understanding, and interpretable evaluations—capabilities that current VLMs have yet to fully realize in nurse training contexts among many others in health care.

To address this challenge, we introduce a VLM system that automatically evaluates nursing skills and provides structured feedback---the first framework applying VLMs to nurse training with curriculum-inspired hierarchical evaluation. Our method mirrors human skill acquisition: progressing from fundamental action recognition to sophisticated error detection and finally to comprehensive procedural reasoning. This systematic design enables assessment across skill levels, from novice to expert practitioners, and supports three principal capabilities: (1) systematic error diagnosis---automatically identifying missing or incorrect subactions in procedures, (2) natural language feedback generation---explaining assessments to learners in actionable terms, and (3) standardized evaluation---ensuring consistent skill competence assessment across institutions. By combining visual and temporal analyses, we enable comprehensive evaluation of procedural execution that transcends individual action recognition.

Our evaluation demonstrates superior performance across multiple nursing procedures, achieving 31.4\% accuracy in procedure identification, 51.7\% relative improvement in fine-grained temporal segmentation (F1: 0.352 at IoU $\geq$ 0.5 vs. base model 0.232), 55.4\% relative improvement in missing action detection (F1: 0.620 vs. base model 0.399), and substantial gains in chronological reasoning, with our best model achieving more than double the baseline performance (Hit@0.5s: 0.1257 vs. baseline 0.0622). These results suggest the framework's effectiveness across all hierarchical assessment tasks. 

In the following, we summarize our contributions to enabling AI-based automated procedural assessment for nursing skills training through innovations in four critical aspects:

\begin{itemize}

    \item \textbf{First VLM-Based Procedure Analysis Framework.} To the best of our knowledge, this work presents the first VLM-based framework for end-to-end procedural analysis, going beyond isolated action recognition.
    
    \item \textbf{Curriculum-Inspired Assessment Hierarchy.} The system mirrors hierarchical human skill acquisition, enabling progressive evaluation from novice to expert levels for various skills.
    
    \item \textbf{Interpretable Natural Language Feedback.} The framework provides interpretable, actionable feedback through contextual explanations, enabling standardized evaluation and skill correction guidance.
    
    \item \textbf{Robust Evaluation across Procedures.} We demonstrate the model's capability to reliably detect procedural errors, identify temporal inconsistencies, and assess missing or out-of-sequence steps across multiple nursing skills, including venipuncture, wound dressings, and urinary retention catheterization, involving complex prescriptive event-based dynamics and potentially significant variation among trainees.

\end{itemize}

Beyond nursing, the developed VLM framework offers a generalizable solution for automatically analyzing prescriptively procedural tasks in various domains with similar characteristics, including medical technician certification, emergency responder training, and precision manufacturing applications. Our approach, integrating sophisticated procedural reasoning with explainable feedback mechanisms, provides a scalable and interpretable foundation for developing procedural assessment technologies in various domains that require workers' procedural precision and reasoning capability.

\section{Background and Related Work}

\subsection{Automated Procedural Assessment: AI Foundations}

Considering the problem of using AI to assess a pre-licensure nursing student's competence on some fundamental nursing skill such as intravenous cannulation or medication administration, current AI systems can detect needle-tissue entry \citep{Ogbonnaya2025AITraining}, track motor movements \citep{Kwon2022AIForMotorLearning}, and identify procedural instruments \citep{Ogbonnaya2025AITraining}. However, they cannot determine whether aseptic technique was maintained, contraindications were recognized, or clinical judgment was applied in alignment with safe nursing practice \citep{Ogbonnaya2025AITraining}. This limitation reveals a fundamental challenge in automated nursing assessment: while AI excels at recognizing what happened, it struggles to evaluate whether it occurred \emph{correctly} according to heuristic nursing standards.

AI methods applied to nursing assessment has progressed from early template-matching systems that performed well in controlled environments but lacked temporal reasoning~\citep{Jain2015Computer}, to deep learning based methods that integrate convolutional and recurrent neural networks capable of modeling procedural workflows with increased spatial-temporal sensitivity~\citep{Khalid2020Deep, Hashemi2025CNN}. Multimodal integration combining visual inputs, pose estimation, and contextual cues has demonstrated improved alignment between AI assessments and expert instructor evaluations~\citep{Kasa2022MultiModal}.

Despite these methodological advances, current AI systems are primarily capable of recognizing individual actions; yet able to assess nursing competency holistically, ranging from clinical reasoning, to patient safety vigilance, to protocol adherence, a spectrum of core components in professional practice~\citep{Javaid2023AutomaticNursingAssessment}. This fundamental limitation emphasizes the need for empowering AI methodology with the capability of sequential reasoning and explainable feedback generation, thus truly achieving comprehensive nursing competency assessment.

\subsection{VLMs for Sequential Activity Understanding}

Recent developments in video-language models (VLMs) have introduced transformative capabilities for temporal reasoning, multi-step analysis, and interpretable procedural understanding. Contemporary VLM architectures, including Video-LLaMA, VideoChatGPT, GPT-4V, and BLIP-2, demonstrate sophisticated integration of visual and linguistic modalities for complex sequential activity analysis~\citep{Zhang2023VideoLLAMA, apollo}.

These models exhibit advanced temporal reasoning capabilities through innovative architectural approaches. Recursive decomposition methods enable breaking complex videos into manageable subtasks with subtask-specific sliding context windows, achieving linear scaling for extended procedural sequences~\citep{fateh2024video}. Token-based temporal modeling, including relative time tokens and continuous temporal token spaces, provides robust representations for fine-grained temporal discrimination, with substantial gains reported in correctness and temporal understanding~\citep{Huang2024LITA}.

Multi-step sequence analysis represents a core strength of modern VLMs, enabled by sophisticated attention mechanisms and memory management strategies. Dynamic memory architectures with adaptive frame selection allow efficient processing of long-form procedural videos, while hierarchical token merging approaches balance global contextual understanding with local feature preservation~\citep{Diko2024ReWind, Weng2024LongVLM}. Chain-of-thought prompting mechanisms enable step-by-step reasoning explanations, providing the interpretable feedback essential for educational applications~\citep{Wang2024SeeDo}.

The interpretability advantages of VLMs stem from their natural language generation capabilities and attention-based explanatory mechanisms. Unlike traditional computer vision approaches that provide opaque confidence scores, VLMs can generate detailed explanations of their reasoning processes, identify critical temporal segments, and provide contextual feedback aligned with human understanding~\citep{Ren2023TimeChat}. These capabilities give hope in addressing fundamental limitations in existing procedural assessment systems regarding explainability and practical utility, which are crucial to nursing education.

While these advances demonstrate VLM capabilities for temporal reasoning and interpretable feedback, their application to nursing education remains unexplored. No existing studies address domain-specific procedural complexity or step-level error detection in clinical contexts. Current implementations target general video understanding rather than specific requirements from nursing skills training, including procedural hierarchies, error taxonomies, and medically-validated feedback. The limited availability of annotated video datasets necessitates alternative approaches such as domain adaptation and few-shot learning  \citep{Cui2023SDA-CLIP, Pachetti2024FewShotMedical}. These gaps establish nurse training as a suitable domain for extending VLM capabilities while addressing practical instructional needs.

\subsection{Nurse Training Assessment: Gaps \& Opportunities}
Nurse training programs worldwide struggle with assessment bottlenecks that threaten educational quality and workforce preparedness. Nursing schools are turning away large numbers of qualified applicants because they don’t have enough faculty, classroom, or clinical site capacity,  ultimately producing fewer nurses than needed in today's healthcare market.  

Instructor-based performance evaluations dominate in the practice of nurse training, yet these evaluations suffer from substantial subjectivity and institutional variability~\citep{gimenes2024protocol, glauberman2023ai}. Evaluator consistency proves problematic, with agreement rates fluctuating dramatically across the assessment of different skills~\citep{cook2022assessing}. Meanwhile, the assessment workload places unsustainable demands on clinical faculty, driving up educational costs~\citep{hewitt2023nursefacultyshortage} while accelerating instructor turnover that weakens program capacity~\citep{smith2023attrition,aldhafeeri2024}. Further, despite established professional standards~\citep{benner2020revisiting,ICN2020competencies}, implementation varies significantly between institutions \citep{Storm2020AssessmentVariability} due to diverse instructional capacity, creating uneven graduate preparedness for practice environments.

While technological solutions promise to address the above assessment challenges, existing AI approaches in nursing evaluation remain inadequate. Most AI systems target isolated gesture recognition or single-action classification~\citep{linardakis2025survey,Tchantchane2023ARO}, failing to capture complete action sequences or clinical decision-making processes~\citep{wang2024lmvideo,Weng2024LongVLM}. This technological limitation creates a fundamental mismatch with nursing education requirements, which demand evaluation of integrated competencies including sequential task execution, safety protocol maintenance, and real-time problem-solving capabilities across extended patient encounters.

The inadequacy of these AI systems stems from nursing procedures presenting distinct analytical challenges that resist conventional video analysis approaches. Considering blood glucose monitoring (BGM), competent performance requires coordinated patient preparation, aseptic technique, proper blood collection, and accurate meter reading—all occurring within fluid temporal boundaries rather than discrete action segments. Automated systems may detect "finger puncture" as an isolated event but cannot evaluate whether proper patient education was provided, contamination risks were minimized throughout, or subsequent care decisions were guided by clinical interpretation of the results.

These limitations reveal that the assessment challenge extends far beyond individual skill recognition to comprehensive procedural reasoning. Competence in nursing skills often requires sustained attention to patient safety, ongoing clinical judgment, and adaptive responses to changing conditions across interconnected care sequences. Cardiopulmonary resuscitation with manual resuscitation bag (CPR-MRB) exemplifies this temporal complexity. The competence demands not only proper chest compression technique and ventilation timing, but sustained assessment of patient responsiveness, continuous monitoring of circulation signs, and adaptive adjustment of intervention intensity~\citep{ILCOR2025EIT}. Surgical hand scrub (SHS) represents an ongoing commitment to segmentation technique that must persist through subsequent patient contact phases~\citep{Yangi2025AIIntegration}. Nasogastric tube (NGT) placement requires continuous evaluation of patient positioning, anatomical considerations, and comfort monitoring throughout extended procedural timeframes~\citep{Haywood2025NGTCompetency}.

In this work, we address the above assessment challenges with AI through a hierarchical learning strategy for automated nursing skill analysis, which sequentially builds the VLM proficiency from fundamental procedure recognition, to granular action segmentation and to advanced diagnostic reasoning.

\begin{figure*}[t]
  \centering
  \includegraphics[width=0.98\textwidth]{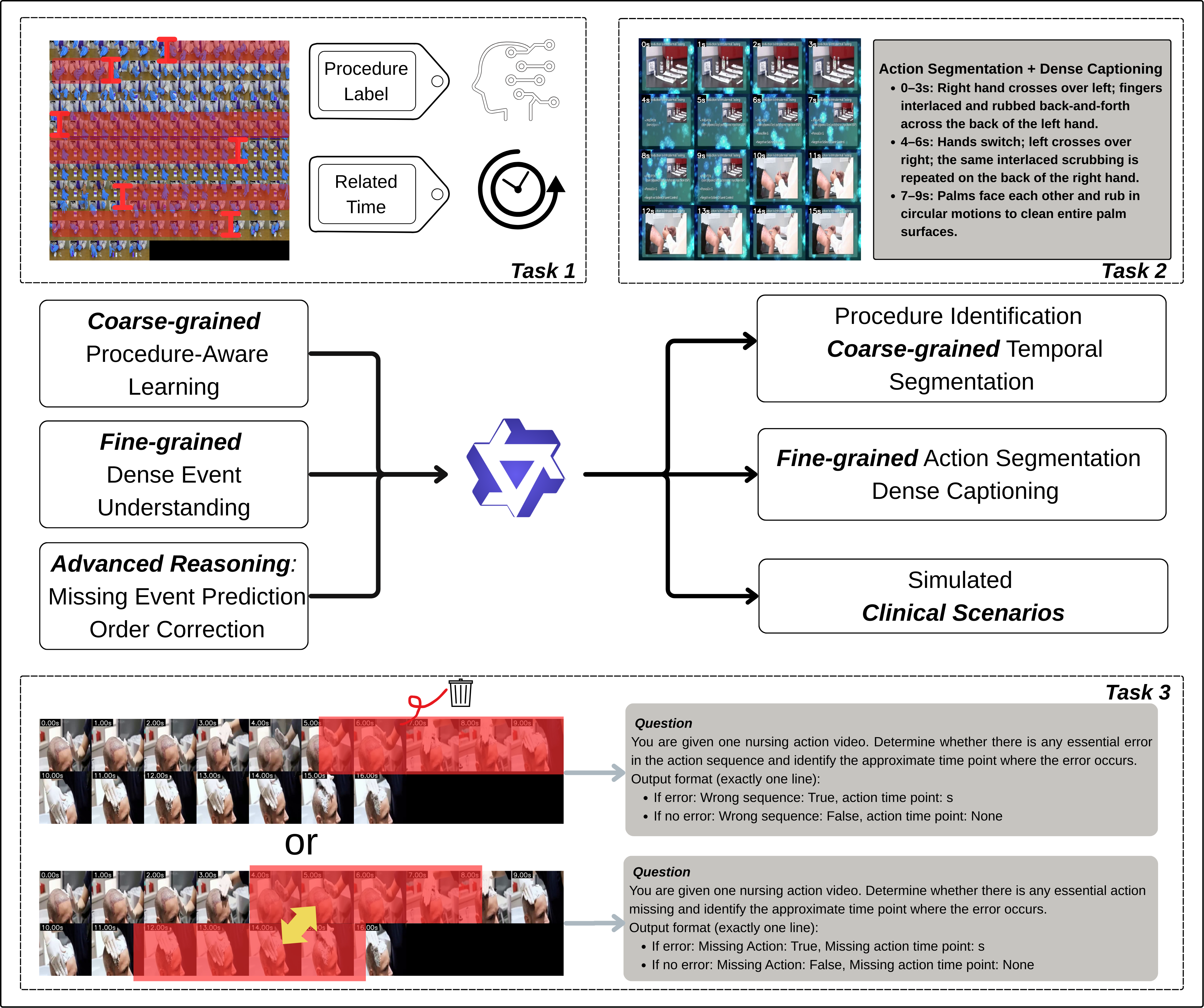} %
  \caption{\textit{Overview of the multistage VLM framework for automated procedural assessment in nursing skills training across three hierarchical tasks.}}
  \label{fig:pipeline}
\end{figure*}
\section{Methodology}
In this section, the development of automated procedural assessment in nursing skills training is demonstrated using a video-language model (VLM) framework guided by curriculum-inspired progressive learning. We begin by outlining the overall research framework, which decomposes the complex procedural assessment challenge into a task hierarchy of three levels. The dataset curation pipeline is then detailed, including (1) the construction of a nursing video corpus with fine-grained subaction annotations, (2) the design of advanced reasoning datasets for masked prediction and shuffled correction, and (3) the visual temporal overlays integration. Next, we present our novel multistage training strategy that systematically spans from coarse-grained procedure-aware learning through fine-grained dense event understanding to higher-order reasoning capabilities, including missing step prediction and shuffled sequence correction. We conclude this section by presenting an experimental validation framework, which assesses the performance of identifying various procedures, the ability of conducting fine-grained action segmentation with dense captioning, and the robustness of the strategy effectiveness under various synthetic procedural error scenarios.

\subsection{VLM-based Framework Overview}
The core objective of this study is to explore the potential and feasibility of leveraging VLMs in nursing skills assessment. To develop the foundational capabilities from ground up, we designed a systematic experimental pipeline aimed at comprehensively evaluating the model performance across diverse training strategies and multimodal scenarios. As illustrated in Figure \ref{fig:pipeline}, the overall framework comprises three hierarchical tasks that progressively assess temporal understanding and reasoning capabilities.

\subparagraph{Task 1: Procedure Identification and Coarse-grained Temporal Segmentation.}
This task aims to evaluate the AI model's capacity to automatically identify the procedures involved in nursing skills and further achieve precise temporal localization of core procedural segments, from completely untrimmed video sequences. Please note that coarse-grained segment sequencing and temporal proximity often imply the preliminary competence of nurses in performing fundamental nursing skills. 
\subparagraph{Task 2: Fine-grained Action Segmentation and Dense Captioning.} 
This task aims to expand the AI model's capacity to offer a richer understanding of nursing action sequences through dense temporal segmentation and detailed stepwise captioning. The model is expected to decompose videos into granular action segments while generating corresponding textual descriptions with temporal alignment. With this task, fine-grained temporal reasoning and vision-language alignment capabilities are critically evaluated as an in-depth competency of nurses. 
\subparagraph{Task 3: Robustness Evaluation on Edited Action Sequences.}
This task aims to evaluate our methodology's capability for causal and chronological reasoning through edited action sequences that are inspired by realistic nursing practice. Specifically, the model must (1) identify procedural incompleteness through missing action identification and infer the missing steps temporal location, and (2) detect temporal sequence errors and determine the specific time slots where procedural misorders occur. To sum up, the above two sub-tasks will help establish the AI model’s efficacy in making causal inference and developing chronological understanding under realistic scenarios.

This multistage approach, from coarse-grained evaluation to fine-grained analysis with dual reasoning assessment, enables systematic evaluation of VLMs’ capabilities on assessing the clinical reasoning competency for nurse training.
\begin{figure*}[t]
  \centering
  \includegraphics[width=0.98\textwidth]{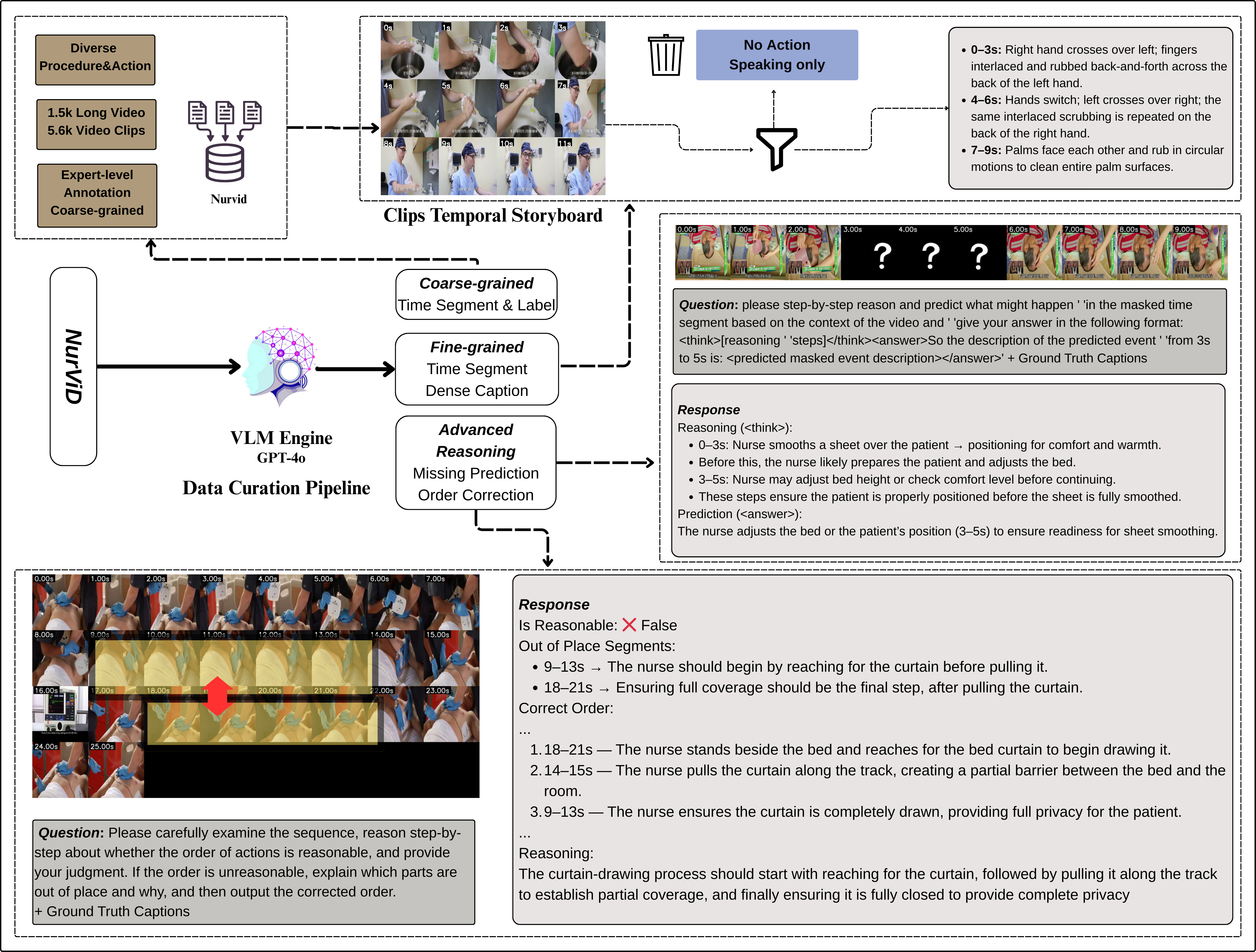} 
  \caption{\textit{VLM-powered data curation workflow producing multi-level annotations for training the proposed AI-based automated procedural assessment model.}}
  \label{fig:dataset}
\end{figure*}

\subsubsection{Nursing Video Corpus}
In this study, all datasets are systematically curated from the recently open-sourced nursing video corpus, \textbf{NurViD}. NurViD constitutes a comprehensive and large-scale multimodal dataset featuring expert-annotated procedural sequences specifically tailored for nursing skill assessment. The corpus encompasses approximately 1.5K video instances spanning 51 distinct nursing procedures and 177 granular action primitives. Compared with existing single-action or limited-scale video benchmarks, NurViD addresses critical scalability limitations and provides a robust foundation for advancing deep learning methodologies in nursing skills assessment.
Each annotation provides precise temporal localization of relevant action segments within extended video sequences, accompanied by corresponding action taxonomies. We leveraged these dense, timestamp-aligned annotations to extract procedural clips for subsequent model training and fine-tuning. The hierarchical annotation structure enables both coarse-grained procedural classification and fine-grained temporal segmentation, supporting comprehensive multimodal instruction-tuning across diverse nursing scenarios.
\subsubsection{Fine-grained Time Segmentation and Dense Caption}
Following the initial curation process, we obtained approximately 5.6K video clips through temporal trimming based on expert annotations. To enable effective training and evaluation of our models on fine-grained action segmentation and dense captioning tasks, a high-quality dataset with comprehensive temporal annotations was essential. However, the original video corpus lacked such granular labels and contained substantial segments with minimal discernible actions or predominantly verbal explanations, rendering them unsuitable for action recognition and temporal reasoning.
To address these limitations, we developed a semi-automated annotation pipeline that simultaneously generated dense temporal annotations and curated high-quality content, leveraging the advanced multimodal understanding capabilities of GPT-4o. At the core of this pipeline is the conversion of video sequences into a static Temporal Storyboard format, as illustrated in Figure~\ref{fig:dataset}. Specifically, frames were systematically sampled at one frame per second (1 FPS), with each frame augmented by timestamp-aligned annotations overlaid in the top-left corner. The timestamped frames were subsequently concatenated horizontally in chronological order to construct a single composite image sequence.
This storyboard representation effectively transforms temporal video dynamics into spatially structured sequences, facilitating direct processing by VLMs through frame-level visual reasoning. The resulting temporal storyboard was input to GPT-4o with comprehensive prompt engineering, instructing the model to function as a domain-expert nursing assessor. The model was tasked with identifying all distinct observable sequential actions and generating precise temporal boundaries (start and end timestamps inferred from visual timestamps), accompanied by concise descriptive captions for each action segment. This integrated approach served two purposes: generating dense temporal annotations for valid procedural clips while simultaneously filtering low-quality content through automated curation. Clips lacking meaningful actions—including static scenes, equipment-only sequences, or predominantly explanatory segments—were systematically excluded from the dataset.
Through this comprehensive annotation-and-curation pipeline, we efficiently constructed a high-quality, action-dense dataset that supported the model training for Stage 2 (i.e., dense event understanding) and established the AI foundation of subsequent reasoning dataset construction and fine-grained temporal analysis.

\subsubsection{Advanced Reasoning: Missing Event Prediction and Sequence Order Correction}
Building on the fine-grained action captioning, we synthesized two additional datasets, specifically designed to enhance the model's capabilities in complex causal and chronological reasoning: missing event prediction and misordered sequence correction. The ground truth for these advanced reasoning tasks were systematically generated through structured prompt engineering with GPT-4o, leveraging the dense captions from our temporally annotated clips.

The \textbf{Missing Event Prediction dataset} was designed to enhance the model's causal inference and contextual reasoning capabilities through temporal event reconstruction. Drawing inspiration from effective infilling techniques, we implemented a masked event paradigm where each annotated video clip underwent selective temporal masking. Specifically, we randomly selected one action segment using its pre-defined temporal boundaries \verb|(start_time, end_time)|, then systematically obscured the corresponding video frames by replacing them with blank sequences.

To generate corresponding training labels, we provided the ground truth captions of remaining visible segments to GPT-4o, accompanied by placeholder indicators for the masked portions. The language model was prompted to generate a structured response containing detailed \verb|reasoning| processes based on visible contextual cues, followed by the \verb|predicted_caption| for the obscured procedural segment. This approach created a comprehensive dataset for training models to logically reconstruct missing events through step-by-step causal explanations.

The \textbf{Sequence Order Correction dataset} was designed to reinforce understanding of standard operating procedures and logical procedural flow inherent to nursing protocols. This dataset employed systematic chronological manipulations of action segments within each procedural clip. We implemented two primary temporal perturbations: \verb|swap| operations, where positions of two randomly selected segments were exchanged, and \verb|shift| transformations, where all segments were advanced sequentially while repositioning the initial segment at the sequence terminus. These manipulations generated procedurally incorrect video sequences that violated established clinical protocols.

Ground truth annotations for this task were generated by providing shuffled action caption sequences to GPT-4o with structured prompt instructions. The model generated comprehensive responses containing four critical components: 1) binary classification determining sequence correctness; 2) identification of misplaced segments with detailed rationales; 3) reconstruction of correct chronological action sequences; and 4) natural language \verb|reasoning| explaining procedural violations (e.g., ``Hand hygiene must be performed before patient contact''). These two synthesized datasets were instrumental for conducting Stage 3 (Missing Event) and Stage 4 (Shuffled Event) training phases, respectively.

\subsubsection{Timestamp Overlay}
To investigate whether explicit temporal cues could enhance the model's temporal reasoning and localization capabilities, we introduced an additional preprocessing step for a subset of experimental conditions: visual timestamp overlay augmentation. The underlying hypothesis posited that providing persistent, explicit temporal signals directly within the visual input stream might improve the model's temporal grounding for time-sensitive procedural tasks. Specifically, we rendered elapsed time information, formatted as \verb|SS:MS|, onto the top-left corner of each video frame, thereby embedding temporal metadata as a visible element within the visual sequence itself. This approach transforms implicit temporal information into explicit visual cues that can be directly processed through the model's vision-language alignment mechanisms.

The rationale for this design is multifaceted. First, it provides unambiguous temporal reference points, potentially reducing the model's reliance on implicitly inferring temporal progression from visual motion cues and procedural context. Second, it reframes aspects of the temporal localization challenge into a text-recognition task, thereby leveraging the optical character recognition (OCR) capabilities inherent in modern VLMs. This dual-modality approach in the design enables the model to integrate both visual procedural understanding and explicit temporal markers for enhanced timestamp-aligned reasoning.

Through this exploratory augmentation strategy, we aimed to examine whether timestamp overlays could facilitate more precise alignment between the model's generated descriptions, predicted segment boundaries, and underlying video timeline. This investigation contributes to comparing the effectiveness of explicit temporal guidance versus implicit temporal inference in nursing procedural assessment. 

\subsection{Multistage Training Strategy}
Our core methodological approach employs a multistage, curriculum-inspired fine-tuning strategy that progressively guides a general-purpose VLM from coarse-grained procedural understanding to fine-grained temporal analysis and to complex causal and chronological reasoning. All training phases utilize the Qwen2.5-VL-7B model (with the 3B version for baseline comparisons).
\subparagraph{Stage 1: Coarse-grained Procedure-Aware Learning}
The primary objective is establishing robust vision-language alignment for fundamental nursing procedure recognition. We adopt a coarse-grained instruction-tuning strategy where the model processes extended nursing video sequences annotated with temporal segment boundaries and corresponding procedural taxonomies as target responses. This stage emphasizes holistic understanding of nursing skills in their entirety, prioritizing broader temporal structure and semantic comprehension over fine-grained step-by-step decomposition. By grounding the model in comprehensive procedure recognition within extended temporal contexts, we establish a foundational understanding that supports more refined learning in subsequent stages.

\subparagraph{Stage 2: Fine-grained Dense Event Understanding}
Upon establishing the procedure-level foundation, the second stage transitions toward fine-grained temporal comprehension and dense captioning. The model undergoes instruction-tuning on densely annotated event segments, pairing temporally trimmed video clips with detailed descriptive captions. With the aim of establishing action relevance achieved, Stage 2 emphasizes precise temporal boundary assignment and comprehensive caption generation describing procedural content and clinical intent. This enhances temporal reasoning capabilities while reinforcing vision-language alignment within the multimodal framework.
\subparagraph{Stage 3: Causal Reasoning through Missing Event Prediction}
Once action-level comprehension is formulated, Stages 3 and 4 transcend isolated video feature recognition toward modeling inter-event correlations. This progression addresses a critical limitation: models trained exclusively on short clips may capture local features effectively but struggle to understand fragment interrelations within broader procedural contexts. These stages emphasize reasoning over temporal dependencies and subaction interconnections.
In Stage 3, we employ missing event prediction to enhance causal reasoning capabilities. The model—pre-trained through Stages 1 and 2—undergoes further instruction-tuning using edited video clips where action segments are intentionally masked. The training objective requires inferring missing events and generating corresponding captions by leveraging contextual information from surrounding visible segments. At this stage of AI model training, we emulate human learners' ability to reconstruct missing procedural steps through reasoning over previously observed actions and their logical continuity. Missing prediction paradigms have demonstrated significant efficacy in enhancing global contextual understanding across deep learning domains. By applying this framework within the video-language domain, we strengthen the model's perception of both local action features and their causal relationships, propelling toward extensive procedural understanding.

\subparagraph{Stage 4: Chronological Reasoning through Sequence Order Correction}
As a complementary approach to further enhance reasoning capabilities, Stage 4 introduces sequence-focused tasks explicitly targeting event order comprehension. The model processes video clips arranged in original chronological order or deliberately perturbed sequences. The learning objective requires determining whether subaction sequences are procedurally correct and contextually reasonable. This stage of AI model training highlights the importance of recognizing individual subactions while reasoning about temporal dependencies and logical progression. By training the model to distinguish coherent procedural flows from disordered sequences, Stage 4 reinforces the capacity to capture underlying narrative structures of nursing procedures. Notably, sequence correction paradigms have been broadly adopted in prior video and language modeling research; here we extend their utility to nursing skills training contexts. This phase strengthens the model’s ability to integrate local observations into consistent global sequences, moving toward a thorough and interpretable understanding of complex procedures.

\subparagraph{Summary:} 
The proposed four-stage curriculum progressively guides the model from coarse-grained recognition toward fine-grained understanding and reasoning. \textbf{Stage 1} establishes foundational alignment through procedure-level recognition on extended videos. \textbf{Stage 2} advances through densely segmented sub-action analysis with descriptive captions, bridging video-text alignment at finer temporal resolution. \textbf{Stage 3} focuses on causal reasoning via missing event prediction, which train the models to enhance contextual dependency inference and procedural continuity reconstruction. \textbf{Stage 4} focuses on chronological reasoning via sequence order correction, which trains the models to distinguish coherent procedural flows from perturbed sequences.

Together, these stages form a coherent, human-inspired learning trajectory: from broad recognition to detailed analysis, culminating in holistic reasoning. This design strengthens the model's capacity to understand nursing procedures across multiple granularities while capturing temporal dependencies and logical structures, thereby advancing video-language modeling towards interpretable applications in clinical education and assessment.

\subsection{Evaluation Metrics}
Our evaluation framework employs two assessment criteria: temporal segment localization accuracy and language quality of generated captions on correctly localized segments. To measure temporal localization performance, we adopt segment-level matching strategies based on temporal Intersection-over-Union (IoU), where predicted segments undergo comparison with ground-truth annotations through greedy one-to-one matching protocols. A predicted segment constitutes a true positive if its IoU with corresponding ground-truth segments exceeds specified thresholds (0.3, 0.5, or 0.7), with Precision, Recall, and F1-scores computed globally across all test sequences. To assess the captioning quality, we compute Rouge-L (measuring longest common subsequence overlap) and Token-level F1 (measuring word-level precision, recall, and harmonic mean) exclusively on matched predicted and ground-truth segment pairs, excluding captions from false positive or false negative segments to ensure metrics specifically reflect description quality where temporal alignment is achieved. This protocol jointly captures segment-level localization accuracy via IoU-based metrics and captioning quality on correctly temporally grounded events. In addition, to evaluate the temporal precision of localized events under varying tolerance levels, we adopt the Hit Ratio metric, commonly used in action localization tasks. A prediction is considered a ``hit'' if the start time of the predicted segment falls within a specified tolerance window of the ground truth start time. We report Hit Ratios at multiple time-window thresholds (0.5s, 1.0s, and 2.0s), which reflect the proportion of correctly localized events under increasingly relaxed error tolerance. This metric complements the IoU-based evaluation by focusing specifically on the accuracy of temporal boundary detection, providing finer-grained insights into the model’s localization performance.

\section{Experiments}

\subsection{Implementation Details}

We conduct all experiments using the Qwen2.5-VL-7B model as our foundational architecture, implementing our multistage training framework on Google Colab A100 GPUs. We train with LoRA fine-tuning (rank 32, alpha 64) targeting all linear layers, with additional fine-tuning of embedding and output head parameters to fully adapt input-output modalities to domain-specific nursing distributions. We utilize DeepSpeed ZeRO-3 optimization with gradient checkpointing for memory efficiency, maintaining a per-device batch size of 1 with gradient accumulation steps of 16 on single GPU, yielding an effective batch size of 16. Our curriculum-inspired training strategy implements learning rates reduced by half at each subsequent stage to ensure stable convergence: Stage 1 employs 1e-5, Stage 2 uses 5e-6, and so forth. All stages utilize cosine learning rate scheduling with 10\% warmup ratio and AdamW optimization. We implement early stopping with a patience of 3 epochs and select the best model based on lowest validation loss across all checkpoints. Training runs up to 10 epochs per stage but may terminate earlier based on convergence criteria. We allocate 10\% of data for evaluation across all stages.

The dataset comprises stage-specific disjoint samples: Stage 1 utilizes 1,457 untrimmed videos for coarse-grained procedure recognition; Stage 2 processes 2,016 temporally segmented clips for fine-grained action understanding; Stage 3 employs 10,454 masked clips for causal reasoning; and Stage 4 uses 3,930 clips (1,953 swap and 1,977 shift operations) for temporal sequence reasoning. Video preprocessing maintains 1 FPS sampling across all stages. We employ mixed-precision training with bfloat16 and implement gradient clipping (max norm=1.0) for training stability.

\subsection{Main Results}
Our experimental evaluation consists of three hierarchical tasks that progressively assess VLM capabilities from coarse-grained procedure recognition to fine-grained temporal understanding and robust clinical reasoning. We evaluate: (1) procedure identification with coarse-grained temporal segmentation on untrimmed nursing videos, (2) fine-grained action segmentation with dense captioning of procedural steps, and (3) robustness evaluation under simulated clinical scenarios including missing and misordered actions. Results demonstrate the effectiveness of our multistage training framework, with each stage contributing distinct capabilities across these evaluation granularities.
\begin{table*}[t!]
\raggedleft
\centering
\caption{Fine-grained action segmentation and dense captioning results sorted by \textbf{F1 at IoU $\geq$ 0.5} in ascending order. P, R, F1 denote Precision, Recall, and F1-score at IoU thresholds 0.3, 0.5, and 0.7. RougeL and TokenF1 evaluate caption quality on correctly localized segments (IoU $\geq$ 0.3). Models with "-ts" use visual timestamp overlays on all stages.}

\small
\resizebox{\textwidth}{!}{%
\begin{tabular}{l|ccc|ccc|ccc|cc}
\hline
\multirow{2}{*}{Model} 
 & \multicolumn{3}{c|}{IoU $\geq$ 0.3} 
 & \multicolumn{3}{c|}{IoU $\geq$ 0.5} 
 & \multicolumn{3}{c|}{IoU $\geq$ 0.7} 
 & \multicolumn{2}{c}{Caption @0.3} \\
 & P & R & F1 & P & R & F1 & P & R & F1 & RougeL & TokenF1 \\
\hline
base-s2-s3-s4-7b-ts   & 0.426 & 0.263 & 0.325 & 0.285 & 0.176 & 0.217 & 0.073 & 0.045 & 0.055 & 0.274 & 0.315 \\
base-7b               & 0.406 & 0.377 & 0.391 & 0.241 & 0.223 & 0.232 & 0.064 & 0.060 & 0.062 & 0.264 & 0.299 \\
base-s2-7b            & 0.429 & 0.385 & 0.406 & 0.279 & 0.250 & 0.264 & 0.092 & 0.083 & 0.087 & 0.307 & 0.344 \\
base-s2-s3-s4-s2-7b-ts& 0.426 & 0.420 & 0.423 & 0.282 & 0.278 & 0.280 & 0.076 & 0.075 & 0.075 & 0.293 & 0.335 \\
base-s3-7b            & 0.442 & 0.449 & 0.445 & 0.280 & 0.284 & 0.282 & 0.093 & 0.094 & 0.094 & 0.274 & 0.311 \\
base-s3-7b-ts         & 0.441 & 0.458 & 0.449 & 0.281 & 0.292 & 0.286 & 0.085 & 0.089 & 0.087 & 0.263 & 0.300 \\
base-s3-s4-7b-ts      & 0.426 & 0.485 & 0.454 & 0.276 & 0.314 & 0.294 & 0.079 & 0.090 & 0.084 & 0.276 & 0.314 \\
base-s3-s4-7b         & 0.397 & 0.541 & 0.458 & 0.263 & 0.358 & 0.303 & 0.072 & 0.098 & 0.083 & 0.276 & 0.316 \\
base-s2-7b-ts         & 0.472 & 0.454 & 0.463 & 0.313 & 0.301 & 0.307 & 0.102 & 0.098 & 0.100 & 0.295 & 0.334 \\
base-s3-s4-s2-7b      & 0.395 & 0.552 & 0.461 & 0.265 & 0.370 & 0.309 & 0.075 & 0.105 & 0.088 & 0.301 & 0.338 \\
base-s3-s4-s2-7b-ts   & 0.413 & 0.524 & 0.462 & 0.281 & 0.357 & 0.315 & 0.076 & 0.097 & 0.085 & 0.297 & 0.333 \\
base-s2-s3-7b-ts      & 0.495 & 0.508 & 0.501 & \textbf{0.347} & \textbf{0.356} & \textbf{0.352} & 0.122 & 0.125 & 0.123 & 0.263 & 0.303 \\
\hline
\end{tabular}}
\label{tab:task2_sorted_asc}
\end{table*}

\begin{table}[!htb]
\centering
\caption{Coarse-grained procedure prediction
 on untrimmed videos. The best performance for each split
 has been highlighted in bold. * denotes the initialization from the model pre-trained on Kinetics 400.}
\begin{tabular}{l|c}
\hline
Baselines & Top-1 Accuracy(\%)\\
\hline
SlowFast & 7.4  \\
C3D & 7.7  \\
I3D& 8.7  \\
VideoMAEv2                                & 24.6 \\
\hline
I3D*                                       & 13.1 \\
SlowFast*                                  & 13.5 \\
C3D*                                       & 14.8\\
Uniformer*                                 & 26.7 \\
VideoMAEv2*                                & 30.1 \\
\hline
base-7b                                    & 5.9\\
base-s1-7b                                 & \textbf{31.4}\\
\hline
\end{tabular}
\label{tab:classification_all}
\end{table}

\subsubsection{Procedure Identification and Coarse-grained Temporal Segmentation}
\begin{table}[tbh!]
\centering
\caption{Performance of base models and Stage-1 trained models on coarse-grained temporal segmentation. Results are reported as F1 scores at IoU thresholds 0.3, 0.5, and 0.7, along with average coverage (Avg. Cov.) and average hit ratio (Avg. Hit).}

\small
\begin{tabular}{l m{2.5em} m{2.5em} m{2.9em}|>{\centering\arraybackslash}m{3.89em}>{\centering\arraybackslash}m{3.35em}}
\hline
Model &F1@0.3 &F1@0.5 &F1@0.7 &Avg. Cov.&Avg. Hit\\
\hline
base-s1-3b & 0.063 & 0.030 & 0.005 & 0.087 & 0.171 \\
base-s1-7b & 0.067 & 0.026 & 0.007 & 0.068 & 0.226 \\
base-3b    & 0.076 & 0.041 & 0.015 & 0.122 & 0.162 \\
base-7b    & \textbf{0.081} & \textbf{0.033} & \textbf{0.011} & 0.138 & 0.206 \\
\hline
\end{tabular}
\label{tab:base-models-f1}
\end{table}

This task evaluates the AI capability in two aspects: procedure identification and temporal segmentation, where the main output consists of procedure names and multiple temporal segments, each defined by start and end timestamps. We compare four model configurations to assess both scale effects and fine-tuning impact: Qwen2.5-VL-3B, Qwen2.5-VL-7B, and their Stage 1-trained variants. As shown in Table~\ref{tab:classification_all}, our base-s1-7b model achieves 31.4\% accuracy, surpassing the previous best baseline VideoMAEv2* (30.1\%) and significantly outperforming traditional models such as I3D (8.7\%) and C3D (7.7\%). The untrained base-7b model achieves only 5.9\% accuracy, highlighting the critical importance of domain-specific instruction tuning. The substantial performance gain from Stage 1 training (25.5\% improvement from 5.9\% to 31.4\%) demonstrates the critical importance of domain-specific instruction tuning for VLMs in procedural recognition tasks. Our trained model achieves competitive performance (31.4\%) compared to specialized video understanding models, with the best baseline VideoMAEv2* reaching 30.1\%. However, as shown in Table \ref{tab:base-models-f1}, Stage 1 training shows limited improvement in temporal boundary detection capabilities, potentially due to video quality constraints. This finding suggests that coarse-grained procedure recognition and fine-grained temporal localization require distinct modeling approaches, motivating our subsequent multistage training strategy.

\subsubsection{Fine-grained Action Segmentation and Dense Captioning}

Evaluation results for temporal segmentation and caption generation across varying IoU thresholds highlight Stage~2’s primary role in enhancing temporal localization. As shown in Table~\ref{tab:task2_sorted_asc}, the best-performing configuration, \textbf{base-s2-s3-7b-ts}, achieves an F1 score of 0.352 at IoU$\geq$0.5, substantially surpassing the base model (0.232), a relative gain of 51.7\%. This confirms Stage~2’s effectiveness for fine-grained temporal understanding, while reasoning stages (S3+S4) yield modest yet consistent gains when applied prior to Stage~2.  

Timestamp overlays (-ts) yield mixed effects across configurations: \textbf{base-s2-7b-ts} improves slightly over \textbf{base-s2-7b} (0.307 vs.\ 0.264), but other variants show less consistent behavior. Captioning quality remains relatively stable (RougeL: 0.263--0.307), suggesting that caption generation is less sensitive to training stage variations than temporal localization. These results demonstrate competent performance on well-structured video segments; however, real-world clinical scenarios often involve incomplete or incorrectly sequenced actions, necessitating more robust reasoning capabilities.

\subsubsection{Robustness Evaluation under Simulated Clinical Scenarios}

\begin{table}[!htb]
\centering
\caption{Missing action detection results sorted by \textbf{F1} (ascending). Precision (P), Recall (R), and F1 measure binary classification of procedural completeness. Hit@0.5s and Hit@1.0s indicate temporal localization accuracy within 0.5 and 1.0 second tolerance windows.}
\begin{tabular}{l>{\centering\arraybackslash}m{1.6em}>{\centering\arraybackslash}m{1.6em}>{\centering\arraybackslash}m{2em}|>{\centering\arraybackslash}m{2.8em}>{\centering\arraybackslash}m{2.8em}}
\hline
Model & P & R & F1 & Hit@0.5s & Hit@1.0s \\
\hline
base-s3-ts        & 0.461 & 0.173 & 0.252 & 0.021 & 0.128 \\
base-s2-s3-ts     & 0.515 & 0.306 & 0.384 & 0.145 & 0.237 \\
base              & 0.503 & 0.331 & 0.399 & 0.066 & 0.198 \\
base-s3-s4-ts     & 0.496 & 0.642 & 0.559 & 0.124 & 0.239 \\
base-s2-ts        & 0.501 & 0.639 & 0.562 & 0.091 & 0.206 \\
base-s2-s3-s4-ts  & 0.538 & 0.635 & 0.582& 0.087 & 0.185 \\
\hline
 base-s3-s4-s2-ts& 0.480& 0.873& \textbf{0.620}& 0.107&0.232\\
\hline
\end{tabular}
\label{tab:task3}
\end{table}

Missing action detection results demonstrate that reasoning stages (S3+S4) combined with dense captioning (S2) achieve optimal performance for identifying procedural incompleteness. Table~\ref{tab:task3} shows that the best-performing model, base-s3-s4-s2-ts, achieves an F1 score of 0.620 with high recall (0.873) but lower precision (0.480), indicating strong capability to identify missing actions while producing some false positives. Compared to the base model (F1: 0.399), this represents a 55.4\% relative improvement.

Models incorporating S3+S4 (base-s3-s4-ts: F1 0.559) or S2 alone (base-s2-ts: F1 0.562) achieve competitive performance individually. However, combining S2 with S3+S4 yields superior results, confirming that both causal reasoning and fine-grained temporal understanding are essential for robust procedural assessment. Temporal localization metrics (Hit@1.0s: 23.2\%) indicate that precise missing action localization remains challenging.

\begin{table}[!htb]
\centering
\caption{Sequence Order Error Detection: Hit Ratios at different thresholds (sorted by Hit@0.5s ascending).}
\begin{tabular}{l|>{\centering\arraybackslash}m{4.5em}>{\centering\arraybackslash}m{4.5em}>{\centering\arraybackslash}m{4em}}
\hline
Model & Hit@0.5s & Hit@1.0s & Hit@2.0s \\
\hline
base              & 0.0622 & 0.4149 & 0.5353 \\
base-s3-ts        & 0.0625 & 0.4375 & 0.5312 \\
base-s3-s4-s2-ts  & 0.0645 & 0.3710 & 0.4785 \\
base-s2-s3-ts     & 0.0676 & 0.4324 & 0.5721 \\
base-s2-ts        & 0.0839 & 0.4650 & 0.5629 \\
base-s3-s4-ts     & 0.0952 & 0.3673 & 0.4694 \\
base-s2-s3-s4-ts  & \textbf{0.1257} & \textbf{0.4431} & \textbf{0.5329} \\
\hline
\end{tabular}
\label{tab:task4_hit_all}
\end{table}

Sequence order error detection results, shown in Table~\ref{tab:task4_hit_all}, highlight distinct contributions of different training stages. The baseline achieves only Hit@0.5s = 0.0622, while adding S2 improves performance substantially to 0.0839, and combining S3 with S4 further boosts performance to 0.0952. In contrast, S3 alone yields limited gains (0.0625), indicating that causal inference alone is less critical for order recovery. The best model (base-s2-s3-s4-ts) reaches 0.1257, nearly doubling the baseline, with consistent improvements across Hit@1.0s and Hit@2.0s. These results suggest stage importance follows S4 $>$ S2 $>$ S3.

\section{Conclusion}

In this work, we establish video-language models as a viable foundation for automated nursing procedural assessment through systematic curriculum-inspired training that mirrors human skill acquisition patterns. Our multistage framework demonstrates that complex clinical competencies can be decomposed into learnable components, with each training phase addressing distinct assessment challenges from basic procedure recognition through fine-grained temporal analysis to advanced error detection and reasoning. The results validate that domain-specific adaptation of general-purpose AI models can effectively bridge the gap between existing video understanding capabilities and specialized healthcare education requirements. While current temporal precision limitations suggest these systems may initially serve as supplementary assessment tools rather than standalone evaluators, the framework provides a scalable methodology for standardizing nursing education across institutions and reducing instructor workload. This research opens pathways for broader applications in procedural skills training across healthcare disciplines, establishing the groundwork for AI-assisted clinical education that can enhance training quality and ultimately contribute to improved patient safety through more consistent and comprehensive competency assessment.

\def\UrlBreaks{\do\/\do-}  

\bibliography{IISE-Trans.bib}
\bibliographystyle{aaai}

\newpage
\appendix
\onecolumn

\end{document}